# Dynamic Local Search for the Maximum Clique Problem

**Wayne Pullan**      W.PULLAN@GRIFFITH.EDU.AU
*School of Information and Communication Technology,*
*Griffith University,*
*Gold Coast, QLD, Australia*

**Holger H. Hoos**      HOOS@CS.UBC.CA
*Department of Computer Science*
*University of British Columbia*
*2366 Main Mall, Vancouver, BC, V6T 1Z4 Canada*

## Abstract

In this paper, we introduce DLS-MC, a new stochastic local search algorithm for the maximum clique problem. DLS-MC alternates between phases of iterative improvement, during which suitable vertices are added to the current clique, and plateau search, during which vertices of the current clique are swapped with vertices not contained in the current clique. The selection of vertices is solely based on vertex penalties that are dynamically adjusted during the search, and a perturbation mechanism is used to overcome search stagnation. The behaviour of DLS-MC is controlled by a single parameter, penalty delay, which controls the frequency at which vertex penalties are reduced. We show empirically that DLS-MC achieves substantial performance improvements over state-of-the-art algorithms for the maximum clique problem over a large range of the commonly used DIMACS benchmark instances.

## 1. Introduction

The maximum clique problem (MAX-CLIQUE) calls for finding the maximum sized subgraph of pairwise adjacent vertices in a given graph. MAX-CLIQUE is a prominent combinatorial optimisation problem with many applications, for example, information retrieval, experimental design, signal transmission and computer vision (Balus & Yu, 1986). More recently, applications in bioinformatics have become important (Pevzner & Sze, 2000; Ji, Xu, & Stormo, 2004). The search variant of MAX-CLIQUE can be stated as follows: Given an undirected graph $G = (V, E)$, where $V$ is the set of all vertices and $E$ the set of all edges, find a maximum size clique in $G$, where a clique in $G$ is a subset of vertices, $C \subseteq V$, such that all pairs of vertices in $C$ are connected by an edge, *i.e.*, for all $v, v' \in C$, $\{v, v'\} \in E$, and the size of a clique $C$ is the number of vertices in $C$. MAX-CLIQUE is $\mathcal{NP}$-hard and the associated decision problem is $\mathcal{NP}$-complete (Garey & Johnson, 1979); furthermore, it is inapproximable in the sense that no deterministic polynomial-time algorithm can find cliques of size $|V|^{1-\epsilon}$ for any $\epsilon > 0$, unless $\mathcal{NP} = \mathcal{ZPP}$ (Håstad, 1999).[1] The best polynomial-time approximation algorithm for MAX-CLIQUE achieves an approximation ratio of $O(|V|/(\log |V|)^2)$ (Boppana & Halldórsson, 1992). Therefore, large and hard instances of MAX-CLIQUE are typically solved using heuristic approaches, in particular, greedy construction algorithms and stochastic local search (SLS) algorithms such as simulated annealing, genetic algorithms and tabu search. (For an overview of these and other methods for solving MAX-CLIQUE, see Bomze, Budinich, Pardalos, & Pelillo, 1999.) It may be noted that the maximum clique problem is equivalent to the independent set problem as well as to the minimum vertex cover problem, and any algorithm for MAX-CLIQUE can be directly applied to these equally fundamental and application relevant problems (Bomze et al., 1999).

---

1. $\mathcal{ZPP}$ is the class of problems that can be solved in expected polynomial time by a probabilistic algorithm with zero error probability.

From the recent literature on MAX-CLIQUE algorithms, it seems that, somewhat unsurprisingly, there is no single best algorithm. Although most algorithms have been empirically evaluated on benchmark instances from the Second DIMACS Challenge (Johnson & Trick, 1996), it is quite difficult to compare experimental results between studies, mostly because of differences in the respective experimental protocols and run-time environments. Nevertheless, particularly considering the comparative results reported by Grosso *et al.* (Grosso, Locatelli, & Croce, 2004), it seems that there are five heuristic MAX-CLIQUE algorithms that achieve state-of-the-art performance.

*Reactive Local Search (RLS)* (Battiti & Protasi, 2001) has been derived from Reactive Tabu Search (Battiti & Tecchiolli, 1994), an advanced and general tabu search method that automatically adapts the tabu tenure parameter (which controls the amount of diversification) during the search process; RLS also uses a dynamic restart strategy to provide additional long-term diversification.

*QUALEX-MS* (Busygin, 2002) is a deterministic iterated greedy construction algorithm that uses vertex weights derived from a nonlinear programming formulation of MAX-CLIQUE.

The more recent *Deep Adaptive Greedy Search (DAGS)* algorithm (Grosso et al., 2004) also uses an iterated greedy construction procedure with vertex weights; the weights in DAGS, however, are initialised uniformly and updated after every iteration of the greedy construction procedure. In DAGS, this weighted iterated greedy construction procedure is executed after an iterative improvement phase that permits a limited amount of plateau search. Empirical performance results indicate that DAGS is superior to QUALEX-MS for most of the MAX-CLIQUE instances from the DIMACS benchmark sets, but for some hard instances it does not reach the performance of RLS (Grosso et al., 2004).

The *k-opt* algorithm (Katayama, Hamamoto, & Narihisa, 2004) is based on a conceptually simple variable depth search procedure that uses elementary search steps in which a vertex is added to or removed from the current clique; while there is some evidence that it performs better than RLS on many instances from the DIMACS benchmark sets (Katayama et al., 2004), its performance relative to DAGS is unclear.

Finally, *Edge-AC+LS* (Solnon & Fenet, 2004), a recent ant colony optimisation algorithm for MAX-CLIQUE that uses an elitist subsidiary local search procedure, appears to reach (or exceed) the performance of DAGS and RLS on at least some of the DIMACS instances.

In this work, we introduce a new SLS algorithm for MAX-CLIQUE algorithm dubbed Dynamic Local Search – Max Clique, DLS-MC, which is based on a combination of constructive search and perturbative local search, and makes use of penalty values associated with the vertices of the graph, which are dynamically determined during the search and help the algorithm to avoid search stagnation.

Based on extensive computational experiments, we show that DLS-MC outperforms other state-of-the-art MAX-CLIQUE search algorithms, in particular DAGS, on a broad range of widely studied benchmark instances, and hence represents an improvement in heuristic MAX-CLIQUE solving algorithms. We also present detailed results on the behaviour of DLS-MC and offer insights into the roles of its single parameter and the dynamic vertex penalties. We note that the use of vertex penalties in DLS-MC is inspired by the dynamic weights in DAGS and, more generally, by current state-of-the-art Dynamic Local Search (DLS) algorithms for other well-known combinatorial problems, such as SAT and MAX-SAT (Hutter, Tompkins, & Hoos, 2002; Tompkins & Hoos, 2003; Thornton, Pham, Bain, & Ferreira, 2004; Pullan & Zhao, 2004); for a general introduction to DLS, see also the work of (Hoos & Stützle, 2004). Our results therefore provide further evidence for the effectiveness and broad applicability of this algorithmic approach.

The remainder of this article is structured as follows. We first describe the DLS-MC algorithm and key aspects of its efficient implementation. Next, we present empirical performance results that establish DLS-MC as the new state-of-the-art in heuristic MAX-CLIQUE solving. This is followed by a more detailed investigation of the behaviour of DLS-MC and the factors determining its performance. Finally, we summarise the main contributions of this work, insights gained from our study and outline some directions for future research.

## 2. The DLS-MC Algorithm

Like the DAGS algorithm by Grosso *et al.*, our new DLS-MC algorithm is based on the fundamental idea of augmenting a combination of iterative improvement and plateau search with vertex penalties that are modified during the search. The iterative improvement procedure used by both algorithms is based on a greedy construction mechanism that starts with a trivial clique consisting of a single vertex and successively expands this clique $C$ by adding vertices that are adjacent to all vertices in $C$. When such an expansion is impossible, there may still exist vertices that are connected to all but one of the vertices in $C$. By including such a vertex $v$ in $C$ and removing the single vertex in $C$ not connected to $v$, a new clique with the same number of vertices can be obtained. This type of search is called *plateau search*. It should be noted that after one or more plateau search steps, further expansion of the current clique may become possible; therefore, DLS-MC alternates between phases of expansion and plateau search.

The purpose of vertex penalties is to provide additional diversification to the search process, which otherwise could easily stagnate in situations where the current clique has few or no vertices in common with an optimal solution to a given MAX-CLIQUE instance. Perhaps the most obvious approach for avoiding this kind of search stagnation is to simply restart the constructive search process from a different initial vertex. However, even if there is random (or systematic) variation in the choice of this initial vertex, there is still a risk that the heuristic guidance built into the greedy construction mechanism causes a bias towards a limited set of suboptimal cliques. Therefore, both DAGS and DLS-MC utilise numerical weights associated with the vertices; these weights modulate the heuristic selection function used in the greedy construction procedure in such a way that vertices that repeatedly occur in the cliques obtained from the constructive search process are discouraged from being used in future constructions. Following this intuition, and consistent with the general approach of dynamic local search (DLS), which is based on the same idea, in this paper, we refer to the numerical weights as *vertex penalties*.

Based on these general considerations, the DLS-MC algorithm works as follows (see also the algorithm outline in Figure 1): After picking an initial vertex from the given graph $G$ uniformly at random and setting the current clique $C$ to the set consisting of this single vertex, all vertex penalties are initialised to zero. Then, the search alternates between an iterative improvement phase, during which suitable vertices are repeatedly added to the current clique $C$, and a plateau search phase, in which repeatedly one vertex of $C$ is swapped with a vertex currently not contained in $C$.

The two subsidiary search procedures implementing the iterative improvement and plateau search phases, *expand* and *plateauSearch*, are shown in Figure 2. Note that both, *expand* and *plateauSearch* select the vertex to be added to the current clique $C$ using only the penalties associated with all candidate vertices. In the case of *expand*, the selection is made from the set $N_I(C)$ of all vertices that are connected to all vertices in $C$ by some edge in $G$; we call this set the *improving neighbour set of* $C$. In *plateauSearch*, on the other hand, the vertex to be added to $C$ is selected from the *level neighbour set of* $C$, $N_L(C)$, which comprises the vertices that are connected to all vertices in $C$ except for one vertex, say $v'$, which is subsequently removed from $C$.

Note that both procedures always maintain a current clique $C$; *expand* terminates when the improving neighbour set of $C$ becomes empty, while *plateauSearch* terminates when either $N_I(C)$ is no longer empty or when $N_L(C)$ becomes empty. Also, in order to reduce the incidence of unproductive plateau search phases, DLS-MC implements the plateau search termination condition of (Katayama et al., 2004) by recording the current clique $(C')$ at the start of the plateau search phase and terminating *plateauSearch* when there is no overlap between the recorded clique $C'$ and the current clique $C$.

At the end of the plateau search phase, the vertex penalties are updated by incrementing the penalty values of all vertices in the current clique, $C$, by one. Additionally, every $pd$ penalty value update cycles (where $pd$ is a parameter called *penalty delay*), all non-zero vertex penalties are decremented by one. This latter mechanism prevents penalty values from becoming too large and allows DLS-MC to 'forget' penalty values over time.

After updating the penalties, the current clique is perturbed in one of two ways. If the penalty delay is greater than one, *i.e.*, penalties are only decreased occasionally, the current clique is reduced to the last vertex $v$ that was added to it. Because the removed vertices all have increased penalty values, they are unlikely to be added back into the current clique in the subsequent iterative improvement phase. This is equivalent to restarting the search from $v$. However, as a penalty delay of one corresponds to a behaviour in which penalties are effectively not used at all (since an increase of any vertex penalty is immediately undone), keeping even a single vertex of the current clique $C$ carries a high likelihood of reconstructing $C$ in the subsequent iterative improvement phase. Therefore, to achieve a diversification of the search, when the penalty delay is one, $C$ is perturbed by adding a vertex $v$ that is chosen uniformly at random from the given graph $G$ and removing all vertices from $C$ that are not connected to $v$.

As stated above, the penalty values are used in the selection of a vertex from a given neighbour set $S$. More precisely, the *selectMinPenalty(S)* selects a vertex from $S$ by choosing uniformly at random from the set of vertices in $S$ with minimal penalty values. After a vertex

```
procedure DLS-MC(G, tcs, pd, maxSteps)
    input: graph G = (V, E); integers tcs (target clique size), pd (penalty delay), maxSteps
    output: clique in G of size at least tcs or 'failed'
begin
    numSteps := 0;
    C := {random(V)};
    initPenalties;
    while numSteps < maxSteps do
        (C, v) := expand(G, C);
        if |C| = tcs then return(C); end if
        C' := C;
        (C, v) := plateauSearch(G, C, C');
        while N_I(C) ≠ ∅ do
            (C, v) := expand(G, C);
            if |C| = tcs then return(C); end if
            (C, v) := plateauSearch(G, C, C');
        end while
        updatePenalties(pd);
        if pd > 1 then
            C := {v};
        else
            v := random(V);
            C := C ∪ {v};
            remove all vertices from C that are not connected to v in G;
        end if
    end while
    return('failed');
end
```

Figure 1: Outline of the DLS-MC algorithm; for details, see text.

has been selected from $S$, it becomes unavailable for subsequent selections until penalties have been updated and perturbation has been performed. This prevents the plateau search phase from repeatedly visiting the same clique. Also, as a safeguard to prevent penalty values from becoming too large, vertices with a penalty value greater than 10 are never selected.

In order to implement DLS-MC efficiently, all sets are maintained using two array data structures. The first of these, the vertex list array, contains the vertices that are currently in the set; the second one, the vertex index array, is indexed by vertex number and contains the index of the vertex in the vertex list array (or $-1$, if the vertex is not in the set). All additions to the set are performed by adding to the end of the vertex list array and updating the vertex index array. Deletions from the set are performed by overwriting the vertex list entry of the vertex to be deleted with the last entry in vertex list and then updating the vertex index array. Furthermore, as vertices can only be swapped once into the current clique during the plateau search phase, the intersection between the current clique and the recorded clique can be simply maintained by recording the size of the current clique at the start of the plateau search and decrementing this by one every time a vertex is swapped

```
procedure expand(G, C)
    input: graph G = (V, E); vertex set C ⊆ V (clique)
    output: vertex set C ⊆ V (expanded clique); vertex v (most recently added vertex)
begin
    while N_I(C) ≠ ∅ do
        v := selectMinPenalty(N_I(C));
        C := C ∪ {v};
        numSteps := numSteps + 1;
    end while;
    return((C, v));
end
```

```
procedure plateauSearch(G, C, C')
    input: graph G = (V, E); vertex sets C ⊆ V (clique), C' ⊆ C (recorded clique)
    output: vertex set C ⊆ V (modified clique); vertex v (most recently added vertex)
begin
    while N_I(C) = ∅ and N_L(C) ≠ ∅ and C ∩ C' ≠ ∅ do
        v := selectMinPenalty(N_L(C));
        C := C ∪ {v};
        remove the vertex from C that is not connected to v in G;
        numSteps := numSteps + 1;
    end while;
    return((C, v));
end
```

Figure 2: Subsidiary search procedures of DLS-MC; for details, see text.

into the current clique. Finally, all array elements are accessed using pointers rather than via direct indexing of the array. [2]

Finally, it may be noted that in order to keep the time-complexity of the individual search steps minimal, the selection from the improving and level neighbour sets does not attempt to maximise the size of the set after the respective search step, but rather chooses a vertex with minimal penalty uniformly at random; this is in keeping with the common intuition that, in the context of SLS algorithms, it is often preferable to perform many relatively simple, but efficiently computable search steps rather than fewer complex search steps.

## 3. Empirical Performance Results

In order to evaluate the performance and behaviour of DLS-MC, we performed extensive computational experiments on all MAX-CLIQUE instances from the Second DIMACS Implementation Challenge (1992–1993)[3], which have also been used extensively for benchmarking purposes in the recent literature on MAX-CLIQUE algorithms. The 80 DIMACS MAX-CLIQUE instances were generated from problems in coding theory, fault diagnosis problems, Keller's conjecture on tilings using hypercubes and the Steiner triple problem, in addition to randomly generated graphs and graphs where the maximum clique has been 'hidden' by incorporating low-degree vertices. These problem instances range in size from less than 50 vertices and 1 000 edges to greater than 3 300 vertices and 5 000 000 edges.

---

2. Several of these techniques are based on implementation details of Henry Kautz's highly efficient Walk-SAT code, see `http://www.cs.washington.edu/homes/kautz/walksat`.

3. http://dimacs.rutgers.edu/Challenges/

All experiments for this study were performed on a dedicated 2.2 GHz Pentium IV machine with 512KB L2 cache and 512MB RAM, running Redhat Linux 3.2.2-5 and using the g++ C++ compiler with the '-O2' option. To execute the DIMACS Machine Benchmark[4], this machine required 0.72 CPU seconds for r300.5, 4.47 CPU seconds for r400.5 and 17.44 CPU seconds for r500.5. In the following, unless explicitly stated otherwise, all CPU times refer to our reference machine.

In the following sections, we first present results from a series of experiments that were aimed at providing a detailed assessment of the performance of DLS-MC. Then, we report additional experimental results that facilitate a more direct comparison between DLS-MC and other state-of-the-art MAX-CLIQUE algorithms.

## 3.1 DLS-MC Performance

To evaluate the performance of DLS-MC on the DIMACS benchmark instances, we performed 100 independent runs of it for each instance, using target clique sizes ($tcs$) corresponding to the respective provably optimal clique sizes or, in cases where such provably optimal solutions are unknown, largest known clique sizes. In order to assess the peak performance of DLS-MC, we conducted each such experiment for multiple values of the penalty delay parameter, $pd$, and report the best performance obtained. The behaviour of DLS-MC for suboptimal $pd$ values and the method used to identify the optimal $pd$ value are discussed in Section 4.2. The only remaining parameter of DLS-MC, $maxSteps$, was set to 100 000 000, in order to maximise the probability of reaching the target clique size in every run.

The results from these experiments are displayed in Table 1. For each benchmark instance we show the DLS-MC performance results (averaged over 100 independent runs) for the complete set of 80 DIMACS benchmark instances. Note that DLS-MC finds optimal (or best known) solutions with a success rate of 100% over all 100 runs per instance for 77 of the 80 instances; the only cases where the target clique size was not reached consistently within the alotted maximum number of search steps ($maxSteps$) are:

- C2000.9, where 93 of 100 runs were successful giving a maximum clique size (average clique size, minimum clique size) of 78 (77.93, 77);

- MANN_a81, where 96 of 100 runs obtained cliques of size 1098, while the remaining runs produced cliques of size 1097; and

- MANN_a45, where all runs achieved a maximum clique size of 344.

For these three cases, the reported CPU time statistics are over successful runs only and are shown in parentheses in Table 1. Furthermore, the expected time required by DLS-MC to reach the target clique size is less than 1 CPU second for 67 of the 80 instances, and an expected run-time of more than 10 CPU seconds is only required for 8 of the 13 remaining instances, all of which have at least 800 vertices. Finally, the variation coefficients (std-dev/mean) of the run-time distributions (measured in search steps, in order to overcome inaccuracies inherent to extremely small CPU times) for the instances on which 100% success rate was obtained were found to reach average and maximum values of 0.86 and 1.59, respectively.

---

4. *dmclique*, ftp://dimacs.rutgers.edu in directory /pub/dsj/clique

| Instance | BR | pd | CPU(s) | Steps | Sols. | Instance | BR | pd | CPU(s) | Steps | Sols. |
|---|---|---|---|---|---|---|---|---|---|---|---|
| brock200_1 | 21 | 2 | 0.0182 | 14091 | 2 | sanr200_0.9 | 42 | 2 | 0.0127 | 15739 | 18 |
| brock200_2 | 12* | 2 | 0.0242 | 11875 | 1 | sanr400_0.5 | 13 | 2 | 0.0393 | 9918 | 4 |
| brock200_3 | 15 | 2 | 0.0367 | 21802 | 1 | sanr400_0.7 | 21 | 2 | 0.023 | 8475 | 61 |
| brock200_4 | 17* | 2 | 0.0468 | 30508 | 1 | C1000.9 | 68 | 1 | 4.44 | 1417440 | 70 |
| brock400_1 | 27 | 15 | 2.2299 | 955520 | 1 | C125.9 | 34* | 1 | 0.0001 | 158 | 94 |
| brock400_2 | 29* | 15 | 0.4774 | 205440 | 1 | C2000.5 | 16 | 1 | 0.9697 | 50052 | 93 |
| brock400_3 | 31 | 15 | 0.1758 | 74758 | 1 | C2000.9 | 78 | 1 | (193.224) | (29992770) | (91) |
| brock400_4 | 33* | 15 | 0.0673 | 28936 | 1 | C250.9 | 44* | 1 | 0.0009 | 845 | 85 |
| brock800_1 | 23 | 45 | 56.4971 | 10691276 | 1 | C4000.5 | 18 | 1 | 181.2339 | 5505536 | 93 |
| brock800_2 | 24 | 45 | 15.7335 | 3044775 | 1 | C500.9 | 57 | 1 | 0.1272 | 72828 | 3 |
| brock800_3 | 25 | 45 | 21.9197 | 4264921 | 1 | c-fat200-1 | 12 | 1 | 0.0002 | 24 | 14 |
| brock800_4 | 26 | 45 | 8.8807 | 1731725 | 1 | c-fat200-2 | 24 | 1 | 0.001 | 291 | 1 |
| DSJC1000_5 | 15* | 2 | 0.799 | 91696 | 25 | c-fat200-5 | 58 | 1 | 0.0002 | 118 | 3 |
| DSJC500_5 | 13* | 2 | 0.0138 | 2913 | 42 | c-fat500-1 | 14 | 1 | 0.0004 | 45 | 19 |
| hamming10-2 | 512 | 5 | 0.0008 | 1129 | 2 | c-fat500-10 | 126 | 1 | 0.0015 | 276 | 3 |
| hamming10-4 | 40 | 5 | 0.0089 | 1903 | 100 | c-fat500-2 | 26 | 1 | 0.0004 | 49 | 18 |
| hamming6-2 | 32 | 5 | < ε | 43 | 2 | c-fat500-5 | 64 | 1 | 0.002 | 301 | 3 |
| hamming6-4 | 4 | 5 | < ε | 3 | 83 | gen200_p0.9_44 | 44* | 1 | 0.001 | 1077 | 4 |
| hamming8-2 | 128 | 5 | 0.0003 | 244 | 100 | gen200_p0.9_55 | 55* | 1 | 0.0003 | 369 | 4 |
| hamming8-4 | 16* | 5 | < ε | 31 | 92 | gen400_p0.9_55 | 55 | 1 | 0.0268 | 18455 | 1 |
| johnson16-2-4 | 8 | 5 | < ε | 7 | 100 | gen400_p0.9_65 | 65 | 1 | 0.001 | 716 | 1 |
| johnson32-2-4 | 16 | 5 | < ε | 15 | 100 | gen400_p0.9_75 | 75 | 1 | 0.0005 | 402 | 1 |
| johnson8-2-4 | 4 | 5 | < ε | 3 | 66 | keller4 | 11* | 1 | < ε | 31 | 98 |
| johnson8-4-4 | 14 | 5 | < ε | 21 | 29 | keller5 | 27 | 1 | 0.0201 | 4067 | 100 |
| MANN_a27 | 126* | 3 | 0.0476 | 41976 | 100 | keller6 | 59 | 1 | 170.4829 | 11984412 | 100 |
| MANN_a45 | 345* | 3 | (51.9602) | (16956750) | (100) | p_hat1000-1 | 10 | 1 | 0.0034 | 230 | 82 |
| MANN_a81 | 1099 | 3 | (264.0094) | (27840958) | (96) | p_hat1000-2 | 46 | 1 | 0.0024 | 415 | 87 |
| MANN_a9 | 16 | 3 | < ε | 21 | 99 | p_hat1000-3 | 68 | 1 | 0.0062 | 1579 | 23 |
| san1000 | 15 | 85 | 8.3636 | 521086 | 1 | p_hat1500-1 | 12* | 1 | 2.7064 | 126872 | 1 |
| san200_0.7_1 | 30 | 2 | 0.0029 | 1727 | 1 | p_hat1500-2 | 65 | 1 | 0.0061 | 730 | 90 |
| san200_0.7_2 | 18 | 2 | 0.0684 | 33661 | 2 | p_hat1500-3 | 94 | 1 | 0.0103 | 1828 | 98 |
| san200_0.9_1 | 70 | 2 | 0.0003 | 415 | 1 | p_hat300-1 | 8* | 1 | 0.0007 | 133 | 13 |
| san200_0.9_2 | 60 | 2 | 0.0002 | 347 | 1 | p_hat300-2 | 25* | 1 | 0.0002 | 87 | 42 |
| san200_0.9_3 | 44 | 2 | 0.0015 | 1564 | 1 | p_hat300-3 | 36* | 1 | 0.0007 | 476 | 10 |
| san400_0.5_1 | 13 | 2 | 0.1641 | 26235 | 1 | p_hat500-1 | 9 | 1 | 0.001 | 114 | 48 |
| san400_0.7_1 | 40 | 2 | 0.1088 | 29635 | 1 | p_hat500-2 | 36 | 1 | 0.0005 | 200 | 14 |
| san400_0.7_2 | 30 | 2 | 0.2111 | 57358 | 1 | p_hat500-3 | 50 | 1 | 0.0023 | 1075 | 36 |
| san400_0.7_3 | 22 | 2 | 0.4249 | 113905 | 1 | p_hat700-1 | 11* | 1 | 0.0194 | 1767 | 2 |
| san400_0.9_1 | 100 | 2 | 0.0029 | 1820 | 1 | p_hat700-2 | 44* | 1 | 0.001 | 251 | 72 |
| sanr200_0.7 | 18 | 2 | 0.002 | 1342 | 13 | p_hat700-3 | 62 | 1 | 0.0015 | 525 | 85 |

Table 1: DLS-MC performance results, averaged over 100 independent runs, for the complete set of DIMACS benchmark instances. The maximum known clique size for each instance is shown in the BR column (marked with an asterisk where proven to be optimal); pd is the optimised DLS-MC penalty delay for each instance; CPU(s) is the run-time in CPU seconds, averaged over all successful runs, for each instance. Average CPU times less than 0.0001 seconds are shown as $< \epsilon$; 'Steps' is the number of vertices added to the clique, averaged over all successful runs, for each instance; 'Sols.' is the total number of distinct maximum sized cliques found for each instance. All runs achieved the best known cliques size shown with the exception of: C2000.9, where 93 of 100 runs were successful giving a maximum clique size (average clique size, minimum clique size) of 78(77.93, 77); MANN_a81, where 96 of 100 runs obtained 1098 giving 1098(1097.96, 1097); and MANN_a45, where all runs achieved a maximum clique size of 344.

It may be interesting to note that the time-complexity of search steps in DLS-MC is generally very low. As an indicative example, `brock800_1` with 800 vertices, $207\,505$ edges and a maximum clique size of 23 vertices, DLS-MC performs, on average, $189\,235$ search steps (*i.e.*, additions to the current clique) per CPU second. Generally, the time-complexity of DLS-MC steps increases with the size of the improving ($N_I$) and level ($N_L$) neighbour sets as well as, to a lesser degree, with the maximum clique size. This relationship can be seen from Table 2 which shows, for the (randomly generated) DIMACS C*.9 and brock*_1 instances, how the performance of DLS-MC in terms of search steps per CPU second decreases as the number of vertices (and hence the size of $N_I$, $N_L$) increases.

| Instance | Vertices | Edges | BR | DLS-MC *pd* | Steps / Second |
|----------|----------|-------|-----|-------------|----------------|
| C125.9   | 125      | 6963    | 34 | 1  | 1587399 |
| C250.9   | 250      | 27984   | 44 | 1  | 939966  |
| C500.9   | 500      | 112332  | 57 | 1  | 572553  |
| C1000.9  | 1000     | 450079  | 68 | 1  | 319243  |
| C2000.9  | 2000     | 1799532 | 78 | 1  | 155223  |
| brock200_1 | 200    | 14834   | 21 | 2  | 774231  |
| brock400_1 | 400    | 59723   | 27 | 15 | 428504  |
| brock800_1 | 800    | 207505  | 23 | 45 | 189236  |

Table 2: Average number of DLS-MC search steps per CPU second (on our reference machine) over 100 runs for the DIMACS C*.9 and brock*_1 instances. The 'BR' and 'DLS-MC *pd*' figures from Table 1 are also shown, as these factors have a direct impact on the performance of DLS-MC. That is, as BR increases, the greater the overhead in maintaining the sets within DLS-MC; furthermore, larger *pd* values cause higher overhead for maintaing penalties, because more vertices tend to be penalised. The C*.9 instances are randomly generated with an edge probability of 0.9, while the brock*_1 instances are constructed so as to 'hide' the maximum clique and have considerably lower densities (*i.e.*, average number of edges per vertex). The scaling of the average number of search steps per CPU second performed by DLS-MC on the C*.9 instances only, running on our reference machine, can be approximated as $9 \cdot 10^7 \cdot n^{-0.8266}$, where $n$ is the number of vertices in the given graph (this approximation achieves an $R^2$ value of 0.9941).

A more detailed analysis of DLS-MC's performance in terms of implementation-independent measures of run-time, such as search steps or iteration counts, is beyond the scope of this work, but could yield useful insights in the future.

## 3.2 Comparative Results

The results reported in the previous section demonstrate clearly that DLS-MC achieves excellent performance on the standard DIMACS benchmark instances. However, a com-parative analysis of these results, as compared to the results found in the literature on other state-of-the-art MAX-CLIQUE algorithms, is not a straight-forward task because of differences in:

- **Computing Hardware:** To date, computing hardware has basically been documented in terms of CPU speed which only allows a very basic means of comparison (*i.e.*, by scaling based on the computer CPU speed which, for example, takes no account of other features, such as memory caching, memory size, hardware architecture, *etc.*). Unfortunately, for some algorithms, this was the only realistic option available to us for this comparison.

- **Result Reporting Methodology:** Most empirical results on the performance of MAX-CLIQUE algorithms found in the literature are in the form of statistics on the clique size obtained after a fixed run-time. To conduct performance comparisons on such data, care must be taken to avoid inconclusive situations in which an algorithm $A$ achieves larger clique sizes than another algorithm $B$, but at the cost of higher run-times. It is important to realise that the relative performance of $A$ and $B$ can vary substantially with run-time; while $A$ may reach higher clique sizes than $B$ for relatively short run-times, the opposite could be the case for longer run-times. Finally, seemingly small differences in clique size may in fact represent major differences in performance, since (as in many hard optimisation problems) finding slightly sub-optimal cliques is typically substantially easier than finding maximal cliques. For example, for C2000.9, the average time needed to find a clique of size 77 (with 100% success rate) is 6.419 CPU seconds, whereas reaching the maximum clique size of 78 (with 93% success rate) requires on average (over successful runs only) of 193.224 CPU seconds.

- **Termination Criteria:** Some MAX-CLIQUE algorithms (such as DAGS) do not terminate upon reaching a given target clique size, but will instead run for a given number of search steps or fixed amount of CPU time, even if an optimal clique is encountered early in the search. It would obviously be highly unfair to directly compare published results for such algorithms with those of DLS-MC, which terminates as soon as it finds the user supplied target clique size.

Therefore, to confirm that DLS-MC represents a significant improvement over previous state-of-the-art MAX-CLIQUE algorithms, we conducted further experiments and analyses designed to yield performance results for DLS-MC that can be more directly compared with the results of other MAX-CLIQUE algorithms. In particular, we compared DLS-MC with the following MAX-CLIQUE algorithms: DAGS (Grosso et al., 2004), GRASP (Resende, Feo, & Smith, 1998) (using the results contained in Grosso et al., 2004), k-opt (Katayama et al., 2004), RLS (Battiti & Protasi, 2001), GENE (Marchiori, 2002), ITER (Marchiori, 2002) and QUALEX-MS (Busygin, 2002). To rank the performance of MAX-CLIQUE algorithms and to determine the dominant algorithm for each of our benchmark instances, we used a set of criteria that are based, primarily, on the quality of the solution and then, when this is deemed equivalent, on the CPU time requirements of the algorithms. These criteria are shown, in order of application, in Table 3.

1. If an algorithm is the only algorithm to find the largest known maximum clique for an instance then it is ranked as the dominant algorithm for that instance.

2. If more than one algorithm achieves a 100% success rate for an instance then the algorithm with the lowest average (scaled) CPU time becomes the dominant algorithm for that instance.

3. If a single algorithm achieves a 100% success rate for an instance then that algorithm becomes the dominant algorithm for that instance.

4. If no algorithm achieves a 100% success rate for an instance, then the algorithm that achieves the largest size clique, has the highest average clique size and the lowest average CPU time becomes the dominant algorithm for that instance.

5. If, for any instance, no algorithm meets any of the four criteria listed above, then no conclusion can be drawn about which is the dominant algorithm for that instance.

Table 3: The criteria used for ranking MAX-CLIQUE algorithms.

| Instance | DLS-MC | | DAGS | | GRASP | |
|---|---|---|---|---|---|---|
| | Clique size | CPU(s) | Clique size | SCPU(s) | Clique size | SCPU(s) |
| brock200_1 | 21 | 0.0182 | 21 | 0.256 | 21 | 4.992 |
| brock200_2 | 12 | 0.0242 | 12 | 0.064 | 12 | 1.408 |
| brock200_3 | 15 | 0.0367 | 15 | 0.064 | 14 | 42.56 |
| brock200_4 | 17 | 0.0468 | 17(16.8,16) | 0.192 | 17 | 3.328 |
| brock400_1 | 27 | 2.2299 | 27(25.35,24) | 1.792 | 25 | 14.976 |
| brock400_2 | 29 | 0.4774 | 29(28.1,24) | 1.792 | 25 | 15.232 |
| brock400_3 | 31 | 0.1758 | 31(30.7,25) | 1.792 | 31(26.2,25) | 14.848 |
| brock400_4 | 33 | 0.0673 | 33 | 1.792 | 25 | 15.232 |
| brock800_1 | 23 | 56.4971 | 23(20.95,20), | 10.624 | 21 | 32 |
| brock800_2 | 24 | 15.7335 | 24(20.8,20) | 10.752 | 21 | 32.96 |
| brock800_3 | 25 | 21.9197 | 25(22.2,21) | 10.88 | 22(21.85,21) | 34.112 |
| brock800_4 | 26 | 8.8807 | 26(22.6,20) | 10.816 | 21 | 33.152 |
| C1000.9 | 68 | 4.44 | 68(65.95,65) | 94.848 | 67(66.1,65) | 154.368 |
| C2000.9 | 78(77.93,77) | 193.224 | 76(75.4,74) | 1167.36 | 75(74.3,73) | 466.368 |
| C4000.5 | 18 | 181.2339 | 18(17.5,17) | 2066.56 | 18(17.75,17) | 466.944 |
| C500.9 | 57 | 0.1272 | 56(55.85,55) | 8.64 | 56 | 80.896 |
| gen200_p0.9_44 | 44 | 0.001 | 44(41.15,40) | 0.576 | 44(41.95,41) | 11.776 |
| gen400_p0.9_55 | 55 | 0.0268 | 53(51.8,51) | 4.608 | 53(52.25,52) | 35.264 |
| gen400_p0.9_65 | 65 | 0.001 | 65(55.4,51) | 4.672 | 65(64.3,63) | 34.56 |
| gen400_p0.9_75 | 75 | 0.0005 | 75(55.2,52) | 4.992 | 74(72.3,69) | 36.16 |
| keller6 | 59 | 170.4829 | 57(56.4,56) | 7888.64 | 55(53.5,53) | 1073.792 |
| MANN_a45 | 344 | 51.9602 | 344(343.95) | 1229.632 | 336(334.5,334) | 301.888 |
| p_hat1000-3 | 68 | 0.0062 | 68(67.85,67) | 71.872 | 68 | 237.568 |
| p_hat1500-1 | 12 | 2.7064 | 12(11.75,11) | 19.904 | 11 | 23.424 |
| san200_0.7_2 | 18 | 0.0684 | 18(17.9,17) | 0.192 | 18(16.55,15) | 3.264 |
| san400_0.7_3 | 22 | 0.4249 | 22(21.7,19) | 1.28 | 21(18.8,17) | 9.856 |
| sanr200_0.9 | 42 | 0.0127 | 42(41.85,41) | 0.576 | 42 | 12.608 |

Table 4: Performance comparison of DLS-MC, DAGS and GRASP for selected DIMACS instances. The SCPU columns contain the scaled DAGS and GRASP average run-times in CPU seconds; DAGS and GRASP results are based on 20 runs per instance, and DLS-MC results are based on 100 runs per instance. In cases where the best known result was not found in all runs, clique size entries are in the format 'maximum clique size (average clique size, minimum clique size)'. DLS-MC is the dominant algorithm for all instances in this table.

Table 4 contrasts performance results for DAGS and GRASP from the literature (Grosso et al., 2004) with the respective performance results for DLS-MC. Since the DAGS and GRASP runs had been performed on a 1.4 GHz Pentium IV CPU, while DLS-MC ran on our 2.2 GHz Pentium IV reference machine, we scaled their CPU times by a factor or 0.64. (Note that this is based on the assumption of a linear scaling of run-time with CPU clock speed; in reality, the speedup is typically significantly smaller.) Using our ranking criteria, this data shows that DLS-MC dominates both DAGS and GRASP for all the benchmark instances listed in Table 4. To confirm this ranking, we modified DAGS so it terminated as soon a given target clique size was reached (this is the termination condition used in DLS-MC) and performed a direct comparison with DLS-MC on all 80 DIMACS instances, running both algorithms on our reference machine. As can be seen from the results of this experiment, shown in Table 5, DLS-MC dominates DAGS on all but one instance (the exception being san1000).

Table 6 shows performance results for DLS-MC as compared to results for k-opt (Katayama et al., 2004), GENE (Marchiori, 2002), ITER (Marchiori, 2002) and RLS (Battiti & Protasi, 2001) from the literature. To roughly compensate for differences in CPU speed, we scaled the CPU times for k-opt, GENE and ITER by a factor of 0.91 (these had been obtained on a 2.0 GHz Pentium IV) and those for RLS (measured on a 450 MHz Pentium II CPU) by 0.21. Using the ranking criteria in Table 3, RLS is the dominant algorithm for instances keller6 and MANN_a45, k-opt is the dominant algorithm for MANN_a81 and DLS-MC is the dominant algorithm, with the exception of C2000.9, for the remainder of the DIMACS instances listed in Table 6. To identify the dominant algorithm for C2000.9, a further experiment was performed, running DLS-MC with its *maxSteps* parameter (which controls the maximum allowable run-time) reduced to the point where the average clique size for DLS-MC just exceeded that reported for RLS. In this experiment, DLS-MC reached the optimum clique size of 78 in 58 of 100 independent runs with an average and minimum clique size of 77.58 and 77, respectively and an average run-time of 85 CPU sec (taking into account all runs). This establishes DLS-MC as dominant over RLS and k-opt on instance C2000.9.

Analagous experiments were performed to directly compare the performance of DLS-MC and k-opt on selected DIMACS benchmark instances; the results, shown in Table 7, confirm that DLS-MC dominates k-opt for these instances.

Finally, Table 8 shows performance results for DLS-MC in comparison with results for QUALEX-MS from the literature (Busygin, 2002); the CPU times for QUALEX-MS have been scaled by a factor of 0.64 to compensate for differences in CPU speed (1.4 GHz Pentium IV CPU *vs* our 2.2 GHz Pentium IV reference machine). Using the ranking criteria in Table 3, QUALEX-MS dominates DLS-MC for instances brock400_1, brock800_1, brock800_2 and brock800_3, while DLS-MC dominates QUALEX-MS for the remaining 76 of the 80 DIMACS instances.

| Instance | DLS-MC Success | DLS-MC CPU(s) | DAGS Success | DAGS CPU(s) | Instance | DLS-MC Success | DLS-MC CPU(s) | DAGS Success | DAGS CPU(s) |
|---|---|---|---|---|---|---|---|---|---|
| brock200_1 | 100 | 0.0182 | 93 | 0.1987 | johnson32-2-4 | 100 | $< \epsilon$ | 100 | 0.0042 |
| brock200_2 | 100 | 0.0242 | 98 | 0.1252 | johnson8-2-4 | 100 | $< \epsilon$ | 100 | $< \epsilon$ |
| brock200_3 | 100 | 0.0367 | 100 | 0.1615 | johnson8-4-4 | 100 | $< \epsilon$ | 100 | 0.0001 |
| brock200_4 | 100 | 0.0468 | 82 | 0.2534 | keller4 | 100 | $< \epsilon$ | 100 | 0.0009 |
| brock400_1 | 100 | 2.2299 | 35 | 3.1418 | keller5 | 100 | 0.0201 | 100 | 0.079 |
| brock400_2 | 100 | 0.4774 | 75 | 2.3596 | keller6 | 100 | 170.4829 | — | — |
| brock400_3 | 100 | 0.1758 | 92 | 2.2429 | MANN_a27 | 100 | 0.0476 | 100 | 0.1886 |
| brock400_4 | 100 | 0.0673 | 99 | 1.653 | MANN_a45 | 100 | 51.9602 | 94 | 8.194 |
| brock800_1 | 100 | 56.4971 | 9 | 20.0102 | MANN_a81 | 96 | 264.0094 | — | — |
| brock800_2 | 100 | 15.7335 | 20 | 18.747 | MANN_a9 | 100 | $< \epsilon$ | 100 | 0.0003 |
| brock800_3 | 100 | 21.9197 | 19 | 19.1276 | p_hat1000-1 | 100 | 0.0034 | 100 | 0.0353 |
| brock800_4 | 100 | 8.8807 | 45 | 16.9227 | p_hat1000-2 | 100 | 0.0024 | 100 | 0.0984 |
| DSJC1000_5 | 100 | 0.799 | 80 | 7.238 | p_hat1000-3 | 100 | 0.0062 | 81 | 37.2 |
| DSJC500_5 | 100 | 0.0138 | 100 | 0.1139 | p_hat1500-1 | 100 | 2.7064 | 69 | 15.609 |
| C1000.9 | 100 | 4.44 | 5 | 2.87 | p_hat1500-2 | 100 | 0.0061 | 100 | 0.4025 |
| C125.9 | 100 | 0.0001 | 100 | 0.0024 | p_hat1500-3 | 100 | 0.0103 | 100 | 6.3255 |
| C2000.9 | 93 | 193.224 | 5 | 2.870608 | p_hat300-1 | 100 | 0.0007 | 100 | 0.0078 |
| C2000.5 | 100 | 0.9697 | 100 | 17.9247 | p_hat300-2 | 100 | 0.0002 | 100 | 0.0033 |
| C250.9 | 100 | 0.0009 | 99 | 0.1725 | p_hat300-3 | 100 | 0.0007 | 100 | 0.0609 |
| C4000.5 | 100 | 181.2339 | — | — | p_hat500-1 | 100 | 0.001 | 100 | 0.0099 |
| C500.9 | 100 | 0.1272 | 4 | 16.2064 | p_hat500-2 | 100 | 0.0005 | 100 | 0.0215 |
| c-fat200-1 | 100 | 0.0002 | 100 | 0.0002 | p_hat500-3 | 100 | 0.0023 | 100 | 0.4236 |
| c-fat200-2 | 100 | 0.001 | 100 | 0.0004 | p_hat700-1 | 100 | 0.0194 | 100 | 0.1217 |
| c-fat200-5 | 100 | 0.0002 | 100 | 0.0012 | p_hat700-2 | 100 | 0.001 | 100 | 0.0415 |
| c-fat500-1 | 100 | 0.0004 | 100 | 0.0005 | p_hat700-3 | 100 | 0.0015 | 100 | 0.1086 |
| c-fat500-10 | 100 | 0.0015 | 100 | 0.0067 | san1000 | 100 | 8.3636 | 100 | 0.967 |
| c-fat500-2 | 100 | 0.0004 | 100 | 0.0009 | san200_0.7_1 | 100 | 0.0029 | 100 | 0.0029 |
| c-fat500-5 | 100 | 0.002 | 100 | 0.0028 | san200_0.7_2 | 100 | 0.0684 | 92 | 0.1001 |
| gen200_p0.9_44 | 100 | 0.001 | 14 | 0.9978 | san200_0.9_1 | 100 | 0.0003 | 100 | 0.0023 |
| gen200_p0.9_55 | 100 | 0.0003 | 100 | 0.0267 | san200_0.9_2 | 100 | 0.0002 | 100 | 0.0368 |
| gen400_p0.9_55 | 100 | 0.0268 | 0 | 9.0372 | san200_0.9_3 | 100 | 0.0015 | 100 | 0.0572 |
| gen400_p0.9_65 | 100 | 0.001 | 27 | 7.1492 | san400_0.5_1 | 100 | 0.1641 | 100 | 0.0336 |
| gen400_p0.9_75 | 100 | 0.0005 | 14 | 8.6018 | san400_0.7_1 | 100 | 0.1088 | 100 | 0.0089 |
| hamming10-2 | 100 | 0.0008 | 100 | 0.1123 | san400_0.7_2 | 100 | 0.2111 | 100 | 0.0402 |
| hamming10-4 | 100 | 0.0089 | 100 | 3.8812 | san400_0.7_3 | 100 | 0.4249 | 90 | 0.5333 |
| hamming6-2 | 100 | $< \epsilon$ | 100 | 0.0003 | san400_0.9_1 | 100 | 0.0029 | 100 | 0.0322 |
| hamming6-4 | 100 | $< \epsilon$ | 100 | $< \epsilon$ | sanr200_0.7 | 100 | 0.002 | 100 | 0.0239 |
| hamming8-2 | 100 | 0.0003 | 100 | 0.0039 | sanr200_0.9 | 100 | 0.0127 | 83 | 0.3745 |
| hamming8-4 | 100 | $< \epsilon$ | 100 | 0.0006 | sanr400_0.5 | 100 | 0.0393 | 93 | 0.231 |
| johnson16-2-4 | 100 | $< \epsilon$ | 100 | 0.0003 | sanr400_0.7 | 100 | 0.023 | 100 | 0.1345 |

Table 5: Success rates and average CPU times for DLS-MC and DAGS (based on 100 runs per instance). For the 80 DIMACS instances, DLS-MC had a superior success rate for 31 instances and, with exception of san1000, required less or the same CPU time than DAGS for all other instances. Entries of '−' signify that the runs were terminated because of excessive CPU time requirements. To obtain a meaningful comparison for DLS-MC and DAGS, for MANN_a45 and MANN_a81, 344 and 1098 respectively were used as best known results in producing this table. For both DLS-MC and DAGS, the average CPU time is over successful runs only. Using the ranking criteria of this study, DAGS is the dominant algorithm for the san1000 instance, while DLS-MC is the dominant algorithm for all other instances.

| | DLS-MC | | k-opt | | RLS | | GENE | ITER |
|---|---|---|---|---|---|---|---|---|
| Instance | Clique size | CPU(s) | Clique size | SCPU(s) | Clique size | SCPU(s) | Avg. Clique size | Avg. Clique size |
| brock200_2 | 12 | 0.0242 | 11 | 0.02184 | 12 | 2.01705 | 10.5 | 10.5 |
| brock200_4 | 17 | 0.0468 | 16 | 0.01911 | 17 | 4.09311 | 15.4 | 15.5 |
| brock400_2 | 29 | 0.4774 | 25(24.6,24) | 0.28028 | 29(26.063,25) | 8.83911 | 22.5 | 23.2 |
| brock400_4 | 33 | 0.0673 | 25 | 0.18291 | 33(32.423,25) | 22.81398 | 23.6 | 23.1 |
| brock800_2 | 24 | 15.7335 | 21(20.8,20) | 2.16034 | 21 | 0.99519 | 19.3 | 19.1 |
| brock800_4 | 26 | 8.8807 | 21(20.5,20) | 2.50796 | 21 | 1.40616 | 18.9 | 19 |
| C1000.9 | 68 | 4.44 | 67 | 6.3063 | 68 | 8.7486 | 61.6 | 61.6 |
| C125.9 | 34 | 0.0001 | 34 | 0.00091 | 34 | 0.00084 | 33.8 | 34 |
| C2000.5 | 16 | 0.9697 | 16 | 13.01846 | 16 | 2.09496 | 14.2 | 14.2 |
| C2000.9 | 78(77.9,77) | 193.224 | 77(75.1,74) | 66.14608 | 78(77.575,77) | 172.90518 | 68.2 | 68.7 |
| C250.9 | 44 | 0.0009 | 44 | 0.05642 | 44 | 0.00609 | 42.8 | 43 |
| C4000.5 | 18 | 181.2339 | 17 | 65.27885 | 18 | 458.44869 | 15.4 | 15.6 |
| C500.9 | 57 | 0.1272 | 57(56.1,56) | 0.82264 | 57 | 0.65604 | 52.2 | 52.7 |
| DSJC1000_5 | 15 | 0.799 | 15 | 5.77941 | 15 | 1.35513 | 13.3 | 13.5 |
| DSJC500_5 | 13 | 0.0138 | 13 | 0.12103 | 13 | 0.04074 | 12.2 | 12.1 |
| gen200_p0.9_44 | 44 | 0.001 | 44 | 0.06643 | 44 | 0.00777 | 39.7 | 39.5 |
| gen200_p0.9_55 | 55 | 0.0003 | 55 | 0.00273 | 55 | 0.00336 | 50.8 | 48.8 |
| gen400_p0.9_55 | 55 | 0.0268 | 53(52.3,51) | 0.56238 | 55 | 0.25284 | 49.7 | 49.1 |
| gen400_p0.9_65 | 65 | 0.001 | 65 | 0.24934 | 65 | 0.0105 | 53.7 | 51.2 |
| gen400_p0.9_75 | 75 | 0.0005 | 75 | 0.16926 | 75 | 0.01071 | 60.2 | 62.7 |
| hamming10-4 | 40 | 0.0089 | 40 | 0.58422 | 40 | 0.01638 | 37.7 | 38.8 |
| hamming8-4 | 16 | < ε | 16 | 0.00182 | 16 | 0.00063 | 16 | 16 |
| keller4 | 11 | < ε | 11 | 0.00091 | 11 | 0.00042 | 11 | 11 |
| keller5 | 27 | 0.0201 | 27 | 0.07371 | 27 | 0.03591 | 26 | 26.3 |
| keller6 | 59 | 170.4829 | 57(55.5,55) | 125.03218 | 59 | 39.86094 | 51.8 | 52.7 |
| MANN_a27 | 126 | 0.0476 | 126 | 0.03276 | 126 | 0.65436 | 125.6 | 126 |
| MANN_a45 | 344 | 51.9602 | 344(343.6,343) | 5.34716 | 345(343.6,343) | 83.7417 | 342.4 | 343.1 |
| MANN_a81 | 1098(1097.96,1097) | 264.0094 | 1099(1098.1,1098) | 84.903 | 1098 | 594.4722 | 1096.3 | 1097 |
| p_hat1500-1 | 12 | 2.7064 | 12 | 15.43997 | 12 | 6.35754 | 10.8 | 10.4 |
| p_hat1500-2 | 65 | 0.0061 | 65 | 0.42224 | 65 | 0.03318 | 63.8 | 63.9 |
| p_hat1500-3 | 94 | 0.0103 | 94 | 2.093 | 94 | 0.04032 | 92.4 | 93 |
| p_hat300-1 | 8 | 0.0007 | 8 | 0.00637 | 8 | 0.00378 | 8 | 8 |
| p_hat300-2 | 25 | 0.0002 | 25 | 0.00546 | 25 | 0.00126 | 25 | 25 |
| p_hat300-3 | 36 | 0.0007 | 36 | 0.0273 | 36 | 0.00441 | 34.6 | 35.1 |
| p_hat700-1 | 11 | 0.0194 | 11 | 0.57876 | 11 | 0.03906 | 9.8 | 9.9 |
| p_hat700-2 | 44 | 0.001 | 44 | 0.04914 | 44 | 0.00588 | 43.5 | 43.6 |
| p_hat700-3 | 62 | 0.0005 | 62 | 0.08008 | 62 | 0.00735 | 60.4 | 61.8 |

Table 6: Performance of DLS-MC, k-opt, RLS, GENE and ITER for selected DIMACS instances. The SCPU columns contain the scaled average run-time in CPU seconds for k-opt and RLS; DLS-MC and RLS results are based on 100 runs per instance, and the k-opt, GENE and ITER results are based on 10 runs per instance. Using the ranking criteria of this study, RLS is the dominant algorithm for instances MANN_a45 and keller6, while DLS-MC is the dominant algorithm for all other instances.

| Instance | DLS-MC Clique size | CPU(s) | k-opt Clique size | SCPU(s) | Instance | DLS-MC Clique size | CPU(s) | k-opt Clique size | SCPU(s) |
|---|---|---|---|---|---|---|---|---|---|
| brock400_2 | 25(24.69,24) | 0.1527 | 25(24.6,24) | 0.280 | C1000.9 | 67(66.07,64) | 0.0373 | 67(66,65) | 6.306 |
| brock400_4 | 25 | 0.0616 | 25 | 0.183 | C2000.9 | 77(75.33,74) | 0.6317 | 77(75.1,74) | 66.146 |
| brock800_2 | 21(20.86,20) | 1.7235 | 21(20.8,20) | 2.160 | C4000.5 | 17 | 1.3005 | 17 | 65.279 |
| brock800_4 | 21(20.65,20) | 1.0058 | 21(20.5,20) | 2.508 | keller6 | 57(55.76,54) | 2.6796 | 57(55.5,55) | 125.032 |

Table 7: Performance of DLS-MC and k-opt where the DLS-MC parameter *maxSteps* has been reduced to the point where the clique size results are comparable to those for k-opt. The CPU(s) values for DLS-MC include the unsuccessful runs; DLS-MC results are based on 100 runs and k-opt results on 10 runs (per instance).

| Instance | DLS-MC Clique size | CPU(s) | QUALEX-MS Clique size | SCPU(s) | Instance | DLS-MC Clique size | CPU(s) | QUALEX-MS Clique size | SCPU(s) |
|---|---|---|---|---|---|---|---|---|---|
| brock200_1 | 21 | 0.0182 | 21 | 0.64 | johnson32-2-4 | 16 | $< \epsilon$ | 16 | 5.12 |
| brock200_2 | 12 | 0.0242 | 12 | < 0.64 | johnson8-2-4 | 4 | $< \epsilon$ | 4 | < 0.64 |
| brock200_3 | 15 | 0.0367 | 15 | 0.64 | johnson8-4-4 | 14 | $< \epsilon$ | 14 | < 0.64 |
| brock200_4 | 17 | 0.0468 | 17 | < 0.64 | keller4 | 11 | $< \epsilon$ | 11 | 0.64 |
| brock400_1 | 27 | 2.2299 | 27 | 1.28 | keller5 | 27 | 0.0201 | 26 | 10.24 |
| brock400_2 | 29 | 0.4774 | 29 | 1.92 | keller6 | 59 | 170.4829 | 53 | 826.24 |
| brock400_3 | 31 | 0.1758 | 31 | 1.28 | MANN_a27 | 126 | 0.0476 | 125 | 0.64 |
| brock400_4 | 33 | 0.0673 | 33 | 1.28 | MANN_a45 | 344 | 51.9602 | 342 | 10.88 |
| brock800_1 | 23 | 56.4971 | 23 | 11.52 | MANN_a81 | 1098(1097.96,1097) | 264.0094 | 1096 | 305.28 |
| brock800_2 | 24 | 15.7335 | 24 | 11.52 | MANN_a9 | 16 | $< \epsilon$ | 16 | < 0.64 |
| brock800_3 | 25 | 21.9197 | 25 | 11.52 | p_hat1000-1 | 10 | 0.0034 | 10 | 17.92 |
| brock800_4 | 26 | 8.8807 | 26 | 11.52 | p_hat1000-2 | 46 | 0.0024 | 45 | 21.76 |
| C1000.9 | 68 | 4.44 | 64 | 17.28 | p_hat1000-3 | 68 | 0.0062 | 65 | 20.48 |
| C125.9 | 34 | 0.0001 | 34 | < 0.64 | p_hat1500-1 | 12 | 2.7064 | 12 | 60.8 |
| C2000.5 | 16 | 0.9697 | 16 | 177.92 | p_hat1500-2 | 65 | 0.0061 | 64 | 71.04 |
| C2000.9 | 78(77.93,77) | 193.224 | 72 | 137.6 | p_hat1500-3 | 94 | 0.0103 | 91 | 69.12 |
| C250.9 | 44 | 0.0009 | 44 | 0.64 | p_hat300-1 | 8 | 0.0007 | 8 | 0.64 |
| C4000.5 | 18 | 181.2339 | 17 | 1500.8 | p_hat300-2 | 25 | 0.0002 | 25 | 0.64 |
| C500.9 | 57 | 0.1272 | 55 | 2.56 | p_hat300-3 | 36 | 0.0007 | 35 | 0.64 |
| c-fat200-1 | 12 | 0.0002 | 12 | < 0.64 | p_hat500-1 | 9 | 0.001 | 9 | 1.92 |
| c-fat200-2 | 24 | 0.001 | 24 | < 0.64 | p_hat500-2 | 36 | 0.0005 | 36 | 2.56 |
| c-fat200-5 | 58 | 0.0002 | 58 | < 0.64 | p_hat500-3 | 50 | 0.0023 | 48 | 2.56 |
| c-fat500-1 | 14 | 0.0004 | 14 | 0.64 | p_hat700-1 | 11 | 0.0194 | 11 | 6.4 |
| c-fat500-10 | 126 | 0.0015 | 126 | 1.28 | p_hat700-2 | 44 | 0.001 | 44 | 7.68 |
| c-fat500-2 | 26 | 0.0004 | 26 | 1.28 | p_hat700-3 | 62 | 0.0015 | 62 | 7.04 |
| c-fat500-5 | 64 | 0.002 | 64 | 1.28 | san1000 | 15 | 8.3636 | 15 | 16.0 |
| DSJC1000_5 | 15 | 0.799 | 14 | 23.04 | san200_0.7_1 | 30 | 0.0029 | 30 | 0.64 |
| DSJC500_5 | 13 | 0.0138 | 13 | 3.2 | san200_0.7_2 | 18 | 0.0684 | 18 | < 0.64 |
| gen200_p0.9_44 | 44 | 0.001 | 42 | < 0.64 | san200_0.9_1 | 70 | 0.0003 | 70 | < 0.64 |
| gen200_p0.9_55 | 55 | 0.0003 | 55 | 0.64 | san200_0.9_2 | 60 | 0.0002 | 60 | 0.64 |
| gen400_p0.9_55 | 55 | 0.0268 | 51 | 1.28 | san200_0.9_3 | 44 | 0.0015 | 40 | < 0.64 |
| gen400_p0.9_65 | 65 | 0.001 | 65 | 1.28 | san400_0.5_1 | 13 | 0.1641 | 13 | 1.28 |
| gen400_p0.9_75 | 75 | 0.0005 | 75 | 1.28 | san400_0.7_1 | 40 | 0.1088 | 40 | 1.92 |
| hamming10-2 | 512 | 0.0008 | 512 | 24.32 | san400_0.7_2 | 30 | 0.2111 | 30 | 1.28 |
| hamming10-4 | 40 | 0.0089 | 36 | 28.8 | san400_0.7_3 | 22 | 0.4249 | 18 | 1.28 |
| hamming6-2 | 32 | $< \epsilon$ | 32 | < 0.64 | san400_0.9_1 | 100 | 0.0029 | 100 | 1.28 |
| hamming6-4 | 4 | $< \epsilon$ | 4 | < 0.64 | sanr200_0.7 | 18 | 0.002 | 18 | 0.64 |
| hamming8-2 | 128 | 0.0003 | 128 | < 0.64 | sanr200_0.9 | 42 | 0.0127 | 41 | < 0.64 |
| hamming8-4 | 16 | $< \epsilon$ | 16 | 0.64 | sanr400_0.5 | 13 | 0.0393 | 13 | 1.28 |
| johnson16-2-4 | 8 | $< \epsilon$ | 8 | < 0.64 | sanr400_0.7 | 21 | 0.023 | 20 | 1.28 |

Table 8: Performance of DLS-MC and QUALEX-MS. The SCPU column contains the scaled run-time for QUALEX-MS in CPU seconds; DLS-MC results are based on 100 runs per instance. Using the ranking criteria of this study, QUALEX-MS is the dominant algorithm for instances brock400_1, brock800_1, brock800_2 and brock800_3, while DLS-MC is the dominant algorithm for all other instances.

Overall, the results from these comparative performance evaluations can be summarised as follows:

- QUALEX-MS is dominant for the brock400_1, brock800_1, brock800_2 and brock800_3 DIMACS instances.

- RLS is the dominant algorithm for the MANN_a45 and keller6 DIMACS instances.

- DAGS is the dominant algorithm for the san1000 DIMACS instance.

- k-opt is the dominant algorithm for the MANN_a81 DIMACS instance.

- DLS-MC is the dominant algorithm for the remaining 72 DIMACS instances.

In addition, within the alotted run-time and number of runs, DLS-MC obtained the current best known results for all DIMACS instances with the exceptions of MANN_a45 and MANN_a81.

## 4. Discussion

To gain a deeper understanding of the run-time behaviour of DLS-MC and the efficacy of its underlying mechanisms, we performed additional empirical analyses. Specifically, we studied the variability in run-time between multiple independent runs of DLS-MC on the same problem instance; the role of the vertex penalties in general and, in particular, the impact of the penalty delay parameter on the performance and behaviour of DLS-MC; and the frequency of pertubation as well as the role of the perturbation mechanism.

These investigations were performed using two DIMACS instances, C1000.9 and brock800_1. These instances were selected because, firstly, they are of reasonable size and difficulty. Secondly, C1000.9 is a randomly generated instance where the vertices in the optimal maximum clique have predominantly higher vertex degree than the average vertex degree (intuitively it would seem reasonable that, for a randomly generated problem, vertices in the optimal maximum clique would tend to have higher vertex degrees). For brock800_1, on the other hand, the vertices in the optimal maximum clique have predominantly lower-than-average vertex degree. (Note that the DIMACS brock instances were created in an attempt to defeat greedy algorithms that used vertex degree for selecting vertices Brockington & Culberson, 1996).

This fundamental difference is further highlighted by the results of a quantitative analysis of the maximum cliques for these instances, which showed that, for C1000.9, averaged over all maximal cliques found by DLS-MC, the average vertex degree of vertices in the maximal cliques is 906 (standard deviation of 9) as compared to 900 (9) when averaged over all vertices; for brock800_1, the corresponding figures were 515 (11) and 519 (13) respectively.

### 4.1 Variability in Run-Time

The variability of run-time between multiple independent runs on a given problem is an important aspect of the behaviour of SLS algorithms such as DLS-MC. Following the methology of Hoos and Stützle (2004), we studied this aspect based on run-time distributions (RTDs) of DLS-MC on our two reference instances.

As can be seen from the empirical RTD graphs shown in Figure 3 (each based on 100 independent runs that all reached the respective best known clique size), DLS-MC shows a large variability in run-time. Closer investigation shows that the RTDs are quite well approximated by exponential distributions (a Kolmogorov-Smirnov goodness-of-fit test failed to reject the null hypothesis that the sampled run-times stem from the exponential distributions shown in the figure at a standard confidence level of $\alpha = 0.05$ with p-values between 0.16 and 0.62). This observation is consistent with similar results for other high-performance SLS algorithms, *e.g.*, for SAT (Hoos & Stützle, 2000) and scheduling problems (Watson, Whitley, & Howe, 2005). As a consequence, performing multiple independent runs of DLS-MC in parallel will result in close-to-optimal parallelisation speedup (Hoos & Stützle, 2004). Similar observation were made for most of the other difficult DIMACS instances.

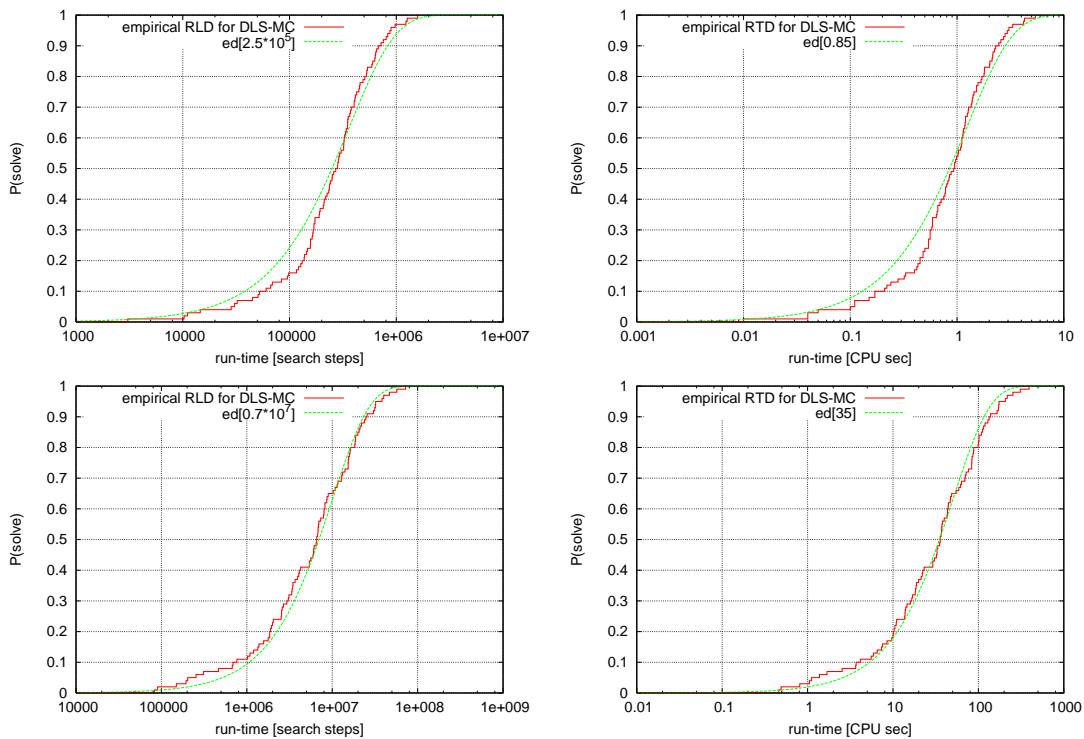

Figure 3: Run-time distributions for DLS-MC applied to C1000.9 (top) and brock800_1 (bottom), measured in search steps (left) and CPU seconds (right) on our reference machine (based on 100 independent runs each of which reached the best known clique size); these empirical RTDs are well approximated by exponential distributions, labelled $ed[m](x) = 1 - 2^{-x/m}$ in the plots.

## 4.2 Penalty Delay Parameter and Vertex Penalties

The penalty delay parameter *pd* specifies the number of penalty increase iterations that must occur in DLS-MC before there is a penalty decrease (by 1) for all vertices that currently have a penalty. For the MAX-CLIQUE problem, *pd* basically provides a mechanism for focusing on lower degree vertices when constructing current cliques. With $pd = 1$ (*i.e.*, no penalties), the frequency with which vertices are in the improving neighbour / level neighbour sets will basically be solely dependent on their degree. Increasing *pd* overcomes this bias towards higher degree vertices, as it allows their penalty values to increase (as they are more often in the current clique), which inhibits their selection for the current clique. This in turn allows lower degree vertices to become part of the current clique. This effect of the penalty delay parameter is illustrated in Figure 4, which shows the correlation between the degree of the vertices and their frequency of being included in the current clique immediately prior to a perturbation being performed within DLS-MC.

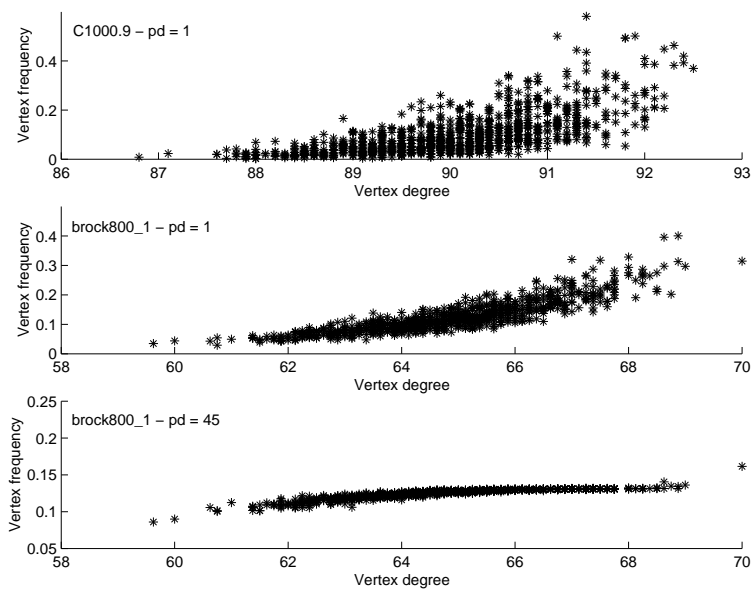

Figure 4: Correlation between the vertex degree and the frequency with which vertices were present in the clique immediately prior to each DLS-MC perturbation. For C1000.9 and brock800_1, with $pd = 1$, the higher degree vertices tend to have a higher frequency of being present in the clique immediately prior to each DLS-MC perturbation. For brock800_1, with $pd = 45$, the frequency of being present in the clique immediately prior to each DLS-MC perturbation is almost independent of the vertex degree.

Currently, *pd* needs to be tuned to a family (or, in the case of the brock instances, a sub-family) of instances. In general, this could be done in a principled way based on RTD graphs, but for DLS-MC, which is reasonably robust with regard to the exact value of the parameter (as shown by Figures 5 and 6), the actual tuning process was a simple, almost interactive process and did not normally require evaluating RTD graphs. Still, fine-tuning based on RTD data could possibly result in further, minor performance improvements.

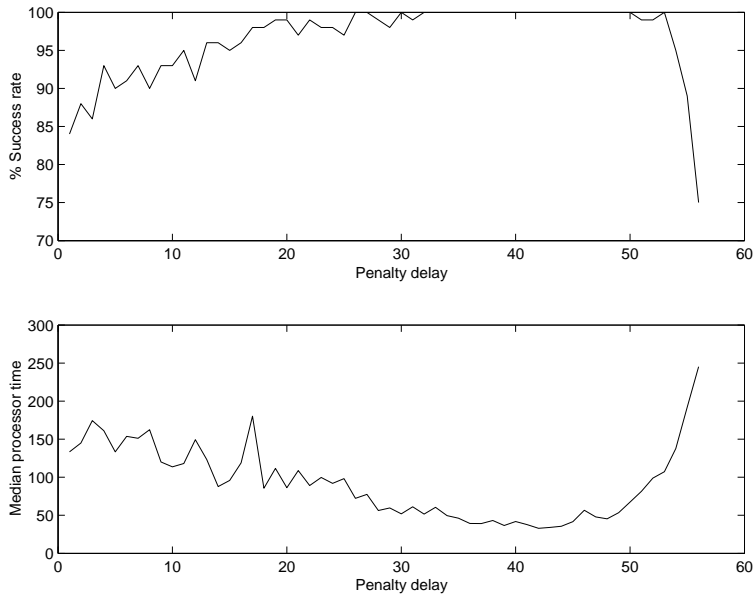

Figure 5: Success rate and median CPU time of DLS-MC as a function of the penalty delay parameter, $pd$, for the benchmark instance brock800_1. Each data point is based on 100 independent runs.

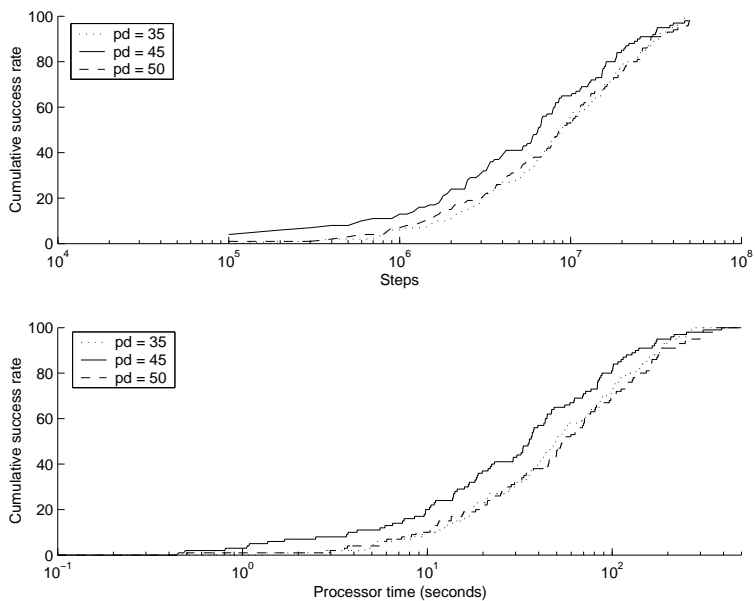

Figure 6: Run-time distributions for DLS-MC on brock800_1 for penalty delays of 35, 45 and 50, measuring run-time in search steps (top) and CPU seconds (bottom). The performance for a penalty delay of 45 clearly dominates that for 35 and 50.

The effect of the penalty delay parameter on the vertex penalties is clearly illustrated in Figure 7, which shows cumulative distributions of the number of penalised vertices at each perturbation in DLS-MC, for representative runs of DLS-MC on the DIMACS brock800_1 instance, for varying values of the parameter *pd*. Note that for brock800_1, the optimal *pd* value of 45 corresponds to the point where, on average, about 90% of the vertices have been penalised. The role of the *pd* parameter is further illustrated in Figure 8, which shows the (sorted) frequency with which vertices were present in the current clique immediately prior to each perturbation for C1000.9 and brock800_1. Note that for both instances, using higher penalty delay settings significantly reduces the bias towards including certain vertices in the current clique. As previously demonstrated, without vertex penalties (*i.e.*, for *pd* = 1), DLS-MC prefers to include high-degree vertices in the current clique, which in the case of problem instances like C1000.9, where optimal cliques tend to consist of vertices with higher-than-average degrees, is an effective strategy. In instances such as brock800_1, however, where the optimal clique contains many vertices of lower-than-average degree, the heuristic bias towards high-degree vertices is misleading and needs to be counteracted, *e.g.*, by means of vertex penalties.

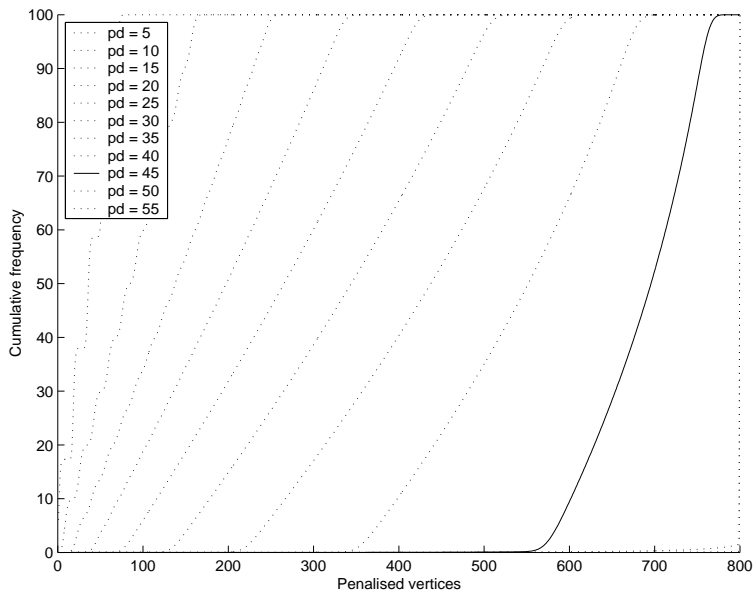

Figure 7: Cumulative distributions of the number of penalised vertices measured at each search perturbation over representative independent runs of DLS-MC on the DI-MACS brock800_1 instance as the penalty delay parameter *pd* is varied (the left most curve corresponds to *pd* = 5). Note that for the approx. optimal penalty delay of *pd* = 45 (solid line), on average about 90% vertices are penalised (*i.e.*, have a penalty value greater than zero).

Generally, by reducing the bias in the cliques visited, vertex penalties help to diversify the search in DLS-MC. At the same time, penalties do not appear to provide a 'learning' mechanism through which DLS-MC identifies those vertices that should be included in the current clique. This is in agreement with recent results for SAPS, a high-performance dynamic local search algorithm for SAT (Hoos & Stützle, 2004).

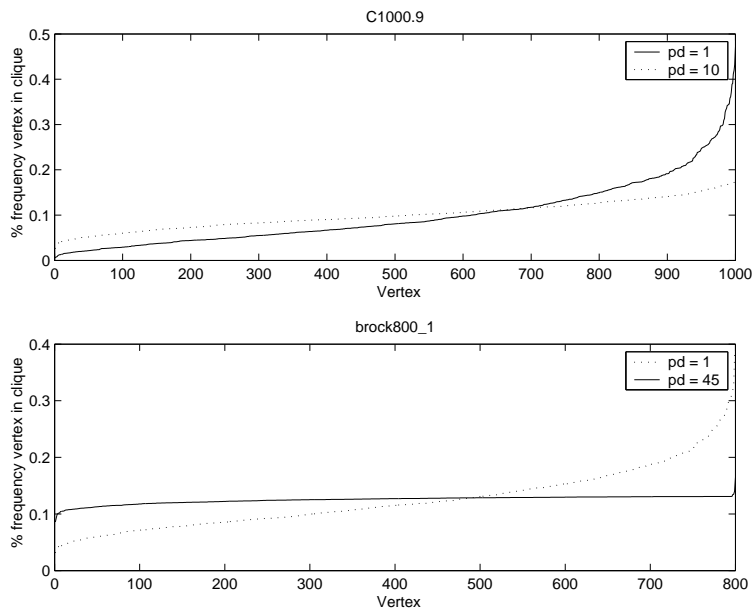

Figure 8: Sorted frequency with which vertices were present in the current clique immediately prior to each DLS-MC perturbation for C1000.9 (top) and brock800_1 (bottom), based on a representative run on each problem instance. Note that by using penalty delay values $pd > 1$, the bias towards using certain vertices more frequently than others is substantially reduced.

## 4.3 Perturbation Mechanism and Search Mobility

To prevent search stagnation, DLS-MC uses a perturbation mechanism that is executed whenever its plateau search procedure has failed to lead to a clique that can be further expanded. Since this mechanism causes major changes in the current clique, it has relatively high time complexity. It is therefore interesting to investigate how frequently these rather costly and disruptive perturbation steps are performed. Figure 9 shows the distribution of the number of improving search steps (*i.e.*, clique expansions) and plateau steps (*i.e.*, vertex swaps) between successive perturbation phases for a representative run of DLS-MC on the C1000.9 instance. Analogous results for brock800_1 are shown in Figure 10. These figures basically show the result of the interactions between the improving and plateau search steps, the perturbation mechanism and the problem structure.

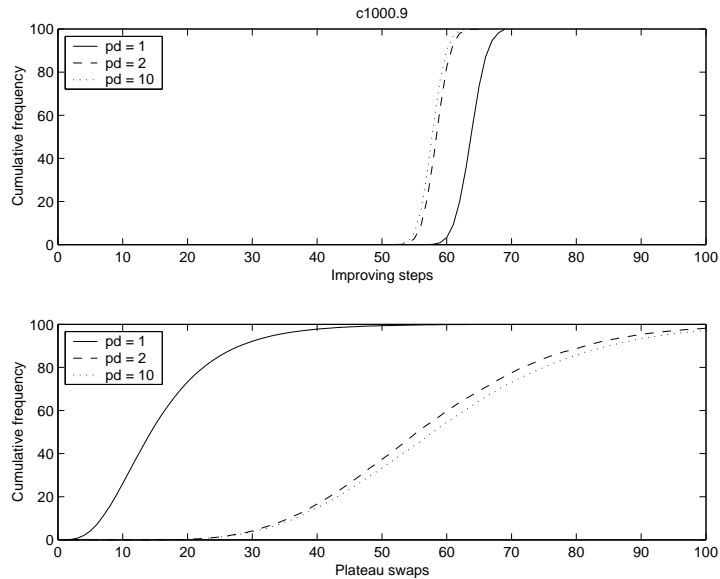

Figure 9: Number of improving search steps and plateau swaps between successive perturbation phases of DLS-MC for C1000.9. The graphs show the cumulative distributions of these measures collected over representative independent runs for each $pd$ value; the solid lines correspond to the approx. optimal penalty delay for this instance, $pd = 1$.

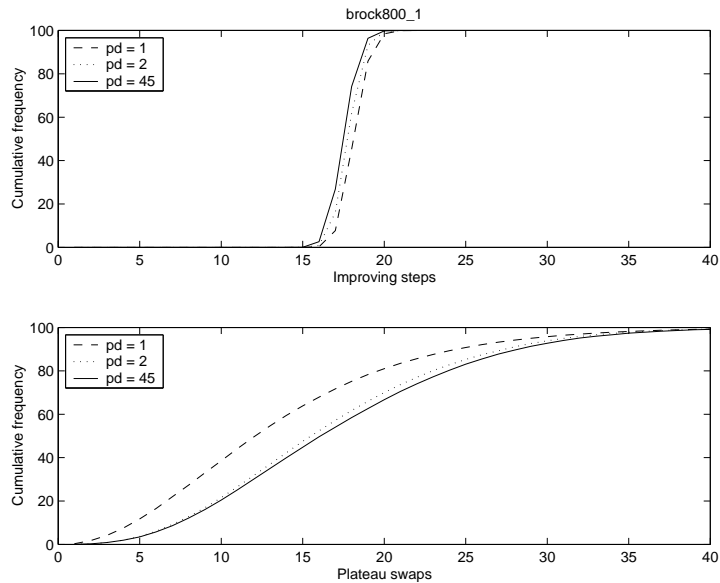

Figure 10: Number of improving search steps and plateau swaps between successive perturbation phases of DLS-MC for brock800_1. The graphs show the cumulative distributions of these measures collected over representative independent runs for each $pd$ value; the solid lines correspond to the approx. optimal penalty delay for this instance, $pd = 45$.

As can be seen from this data, when compared to higher penalty delay values, $pd = 1$ results in significantly shorter plateau phases and somewhat longer improvement phases. At the same time, the differences in the behaviour of DLS-MC observed for various penalty delay values greater than one are relatively small. One explanation for this phenomenon lies in the fact that for $pd = 1$, effectively no vertex penalties are used, and consequently, the selection from the improving and level neighbours sets in each search step is less constrained. Intuitively, this should make it easier to find exits off plateaus in the underlying search landscape and to follow gradients for a larger number of search steps.

Whether this renders the search more efficient clearly depends on the topology of the given search landscape. Instance C1000.9 has at least 70 optimal solutions (see Table 1), and by construction, these optimal cliques have higher-than-average vertex degree. This suggests that the respective search landscape has a relatively high fitness-distance correlation, which would explain why this problem instance is relatively easy to solve and also why using the less radical perturbation mechanism associated with $pd = 1$ (which adds a randomly chosen vertex $v$ to the current clique and removes all vertices not connected to $v$) provides sufficient diversification to the search process. Instance brock800_1, on the other hand, appears to have only a single optimal solution but many near-optimal solutions (i.e., large but non-optimal cliques that cannot be further extended), since by construction, its optimal clique has lower-than-average vertex degree. This suggests that the respective search landscape has relatively low fitness-distance correlation, and therefore, the more radical perturbation mechanism used for $pd > 1$ (which restarts clique construction from the most recently added vertex and uses vertex penalties for diversification) is required in order to obtain good performance; this hypothesis is also in agreement with the relatively high cost for solving this problem instance.

To further investigate the efficacy of perturbation in DLS-MC as a diversification mechanism, we measured the relative mobility of the search, defined as the Hamming distance between the current cliques (i.e., number of different vertices) at consecutive perturbations divided by two times the maximum clique size, for representative runs of DLS-MC on instances C1000.9 and brock800_1 (this mobility measure is closely related to those used in previous studies (Schuurmans & Southey, 2000)). As can be seen from Figure 11, there is a large difference in mobility between the two variants of the perturbation mechanism for $pd = 1$ and $pd > 1$; the former restarts the search from a randomly chosen vertex and consequently leads to a large variability in Hamming distance to the previous clique, while the latter restarts from the most recently added vertex, using vertex penalties to increase search diversification, and hence shows consistently much higher mobility. Note that when vertex penalties are used (i.e., $pd > 1$), the $pd$ value has no significant effect on search mobility. At the same time, as previously observed (see Figure 5), the performance of DLS-MC does significantly depend on the penalty update delay $pd$. This demonstrates that in order to achieve peak performance, the increased mobility afforded by the use of vertex penalties needs to be combined with the correct amount of additional diversification achieved by using a specific penalty update delay.

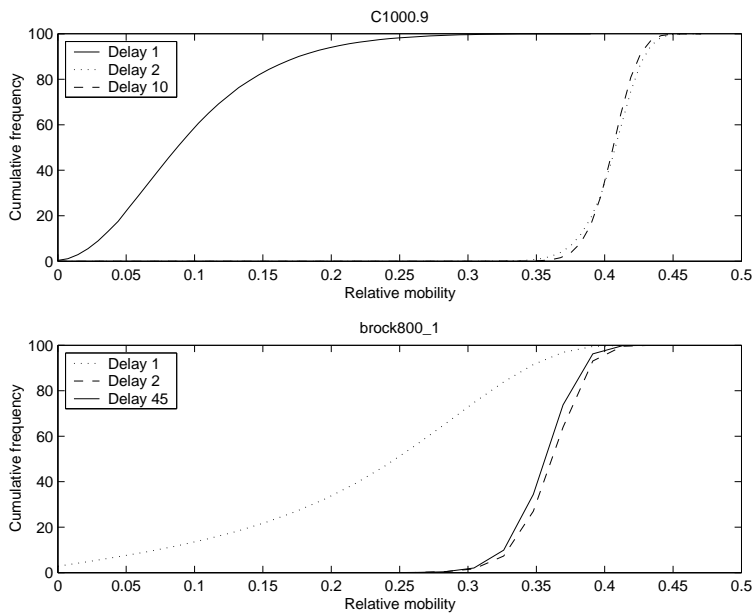

Figure 11: Mobility of search between consecutive perturbation phases in DLS-MC for instances C1000.9 (top) and brock800_1 (bottom). Mobility is measured in terms of relative Hamming distance, *i.e.*, number of different vertices between the respective cliques divided by two times the maximum clique size. The graphs show the cumulative distributions of relative mobility measurements collected over representative independent runs for each *pd* value and problem instance; the solid lines correspond to the respective approx. optimal *pd* values.

## 5. Conclusions and Future Work

We have demonstrated how by applying the general paradigm of dynamic local search to the maximum clique problem, the state of the art in MAX-CLIQUE solving can be improved. Our new algorithm, DLS-MC, has some similarity to previous MAX-CLIQUE algorithms, in particular to the recently introduced DAGS algorithm: Both algorithms use vertex penalties to guide the heuristic selection of vertices when searching for maximum cliques. However, unlike DAGS, which has an initial phase of unweighted greedy construction search, DLS-MC uses and updates the vertex penalties throughout the entire search process. Furthermore, weight updates in DAGS are monotone while, in DLS-MC, vertex penalties are subject to increases as well as to occasional decreases, which effectively allows the algorithm to 'forget' vertex penalties over time. Furthermore, DLS-MC selects the vertex to be added to the current clique in each step solely based on its penalty, while vertex selection in DAGS is based on the total weight of the neighbouring vertices and hence implicitly uses vertex degree for heuristic guidance. The fact that DLS-MC, although conceptually slightly simpler, outperforms DAGS on all but one of the standard DIMACS benchmark instances in combination with its excellent performance compared to other high-performance MAX-CLIQUE algorithms clearly demonstrates the value of the underlying paradigm of dynamic local search with non-monotone penalty dynamics.

The work presented in this article can be extended in several directions. In particular, it would be interesting to investigate to which extent the use of multiplicative penalty update mechanisms in DLS-MC instead of its current additive mechanism can lead to further performance improvements. We also believe that the current implementation of DLS-MC can be further optimised. For example, for each selection of a vertex to be added to the current clique, our implementation of DLS-MC performs a complete scan of either the improving or plateaus sets to build the list of vertices with the lowest penalties; it would probably be more efficient to maintain this list by means of an incremental update scheme. Another very interesting direction for future research is to develop mechanisms for automatically adjusting DLS-MC's penalty delay parameter during the search, similar to the scheme used for dynamically adapting the tabu tenure parameter in RLS (Battiti & Protasi, 2001) and Reactive Tabu Search (Battiti & Tecchiolli, 1994), or the mechanism used for DLS-MC for controlling the noise parameter in Adaptive Novelty$^+$ (Hoos, 2002). Finally, given the excellent performance of DLS-MC on standard MAX-CLIQUE instances reported here suggests that the underlying dynamic local search method has substantial potential to provide the basis for high-performance algorithms for other combinatorial optimisation problems, particularly weighted versions of MAX-CLIQUE and conceptually related clustering problems.

## Acknowledgments

The authors would like to thank Liang Zhao for her participation in performing some of the initial experiments for this paper.